\documentclass[10pt,twocolumn,letterpaper]{article}

\usepackage{cvpr}
\usepackage{times}
\usepackage{epsfig}
\usepackage{graphicx}
\usepackage{amsmath}
\usepackage{amssymb}

\usepackage[pagebackref=true,breaklinks=true,citecolor=blue,letterpaper=true,colorlinks,bookmarks=false]{hyperref}

\cvprfinalcopy 

\def\cvprPaperID{465} 

\usepackage[square,numbers,sort&compress]{natbib}
\usepackage{algpseudocode}
\usepackage{multirow}
\usepackage{booktabs}
\usepackage{algorithm}
\usepackage{caption}
\usepackage{subcaption}
\usepackage[normalem]{ulem}
\usepackage{xparse}

\newcommand{\prop}[1]{\small\tt{#1}}
\NewDocumentCommand{\rot}{O{45} O{1em} m}{\makebox[#2][l]{\rotatebox{#1}{#3}}}%

\ifcvprfinal\fi
\begin{document}

\title{Seeing through the Human Reporting Bias: \\ Visual Classifiers from Noisy Human-Centric Labels}

\author{	
    Ishan Misra$^{1}$ \thanks{Work done during internship at Microsoft Research.}  \quad \quad C. Lawrence Zitnick$^3$ \quad \quad Margaret Mitchell$^2$ \quad \quad Ross Girshick$^3$ \\    
    $^1$Carnegie Mellon University \quad $^2$Microsoft Research \quad $^3$Facebook AI Research
}

\maketitle

\begin{abstract}
When human annotators are given a choice about what to label in an image, they apply their own subjective judgments on what to ignore and what to mention.  We refer to these noisy ``human-centric'' annotations as exhibiting human reporting bias.  Examples of such annotations include image tags and keywords found on photo sharing sites, or in datasets containing image captions. In this paper, we use these noisy annotations for learning visually correct image classifiers. Such annotations do not use consistent vocabulary, and miss a significant amount of the information present in an image; however, we demonstrate that the noise in these annotations exhibits structure and can be modeled. We propose an algorithm to decouple the human reporting bias from the correct visually grounded labels. Our results are highly interpretable for reporting ``what's in the image'' versus ``what's worth saying.''  We demonstrate the algorithm's efficacy along a variety of metrics and datasets, including MS COCO and Yahoo Flickr 100M. We show significant improvements over traditional algorithms for both image classification and image captioning, doubling the performance of existing methods in some cases.

\end{abstract}

\section{Introduction}
\begin{figure}[!t]
\centering
\includegraphics[width=0.48\textwidth]{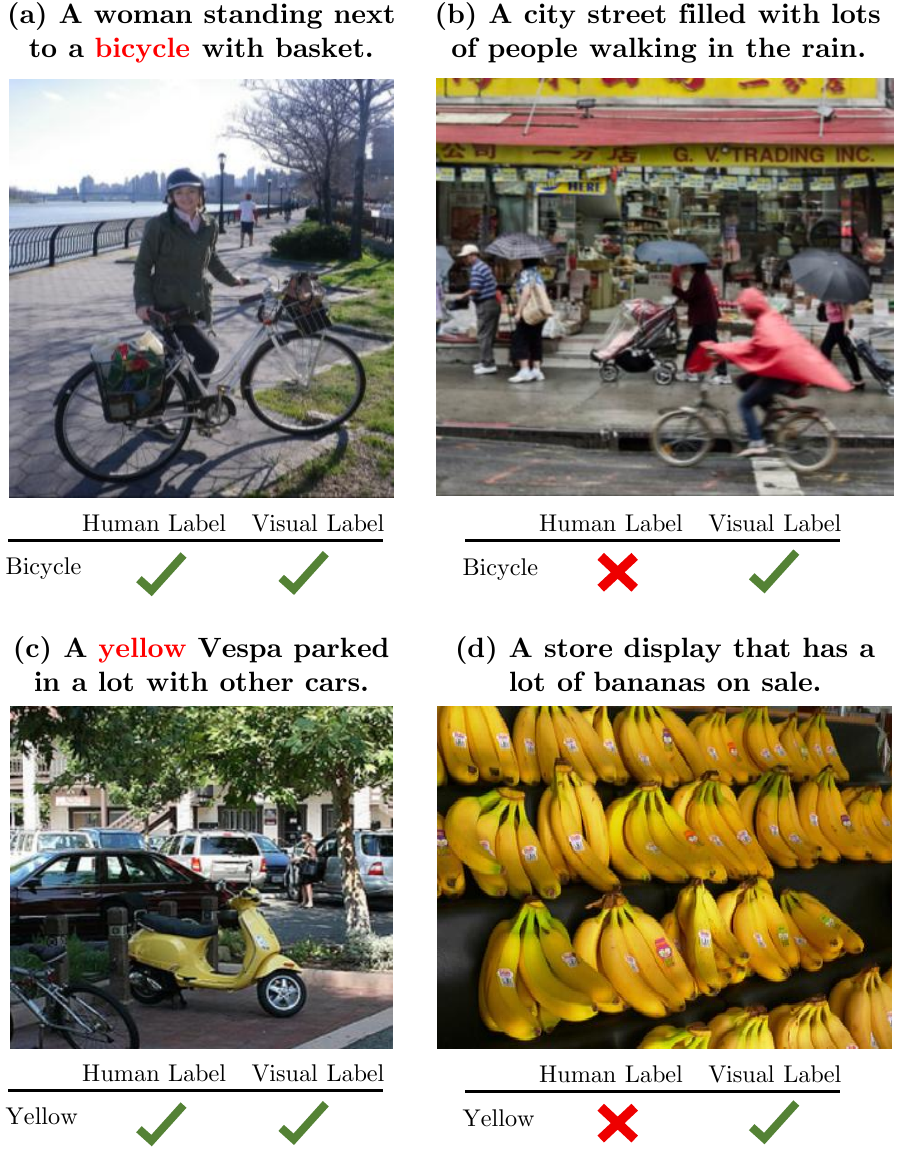}
\vspace{-0.05in}
\caption{Human descriptions capture only some of the visual concepts present in an image. For instance, the bicycle in (a) is described, while the bicycle in (b) is not mentioned. The Vespa in (c) is described as ``yellow'', while the bananas in (d) are not, as being yellow is typical for bananas.}
\vspace{-0.1in}
\label{fig:teaser}
\end{figure}

Visual concept recognition is a fundamental computer vision task with a broad range of applications in science, medicine, and industry.
Supervised learning of visual concept classifiers has been highly successful partly due to the use of large-scale, high-quality datasets (\eg, \cite{PASCAL,COCO,ImageNet}).
Depending on the complexity of the supported task, these datasets generally include annotations for 100s to 1000s of `typical' concepts.
To support an even broader range of applications, it is necessary to train classifiers for tens or even hundreds of thousands of visual concepts that may not be typical.
Since supervised learning methods require exhaustive and clean annotations, one would require high quality datasets with orders of magnitude more annotations to train such methods. However, creating such datasets is expensive.
An alternative approach is to relax this requirement of pristinely labeled data. The learning algorithm can be enabled to use readily-available sources of annotated data, such as user-generated image tags or captions from social media services like Flickr or Instagram. Such datasets easily scale to hundreds of millions of photos with hundreds of thousands of distinct tags \cite{yfcc100m}.

Images annotated with human-written tags \cite{yfcc100m} or captions \cite{COCOCaption} focus on the most important or salient information in an image, as judged implicitly by the annotator. These annotations lack information on minor objects or information that may be deemed unimportant, a phenomenon known as \emph{reporting bias} \cite{gordon2013reportingBias}.
For example, Figure \ref{fig:teaser} illustrates two concepts ({\prop bicycle}, {\prop yellow}) that are each present in two images, but only mentioned in one.
The bicycle may be considered irrelevant to the overall image in (b); and the bananas in (d) are not described as yellow because humans often omit an object's typical properties when referring to it \cite{MitchellEtAl13,WesterbeekEtAl15}.
Following \cite{berg2012understanding}, we refer to this type of labeling as \emph{human-centric annotation}.

Training directly on human-centric annotations does not yield a credible visual concept classifier.
Instead, it leads to a classifier that attempts to \emph{mimic} the reporting bias of the annotators.
To separate reporting bias from visual ground truth, we propose to train a model that explicitly factors human-centric label prediction into a \emph{visual presence} classifier (\ie, ``Is this concept visually present in this image?'') and a \emph{relevance} classifier (\ie, ``Is this concept worth mentioning in this image, given its visual presence?''). We train all these classifiers jointly and end-to-end as multiple ``heads'' branching from the same shared convolutional neural network (ConvNet) trunk \cite{lecun1989backpropagation,VGG}.

We demonstrate improved performance on several tasks and datasets. Our experiments on the MS COCO Captions dataset \cite{COCOCaption} show an improvement in mean average precision (mAP) for the learned visual classifiers when evaluated on both fully labeled data (using annotations from the MS COCO detection benchmark~\cite{COCO}) and on the human generated caption data. We also show that using such visual predictions improves image caption generation quality. Our results on the Yahoo Flickr 100M dataset~\cite{yfcc100m} demonstrate the ability of our model to learn from ``in the wild'' data (noisy Flickr tags) and \textbf{double the performance} of the baseline classification model. Apart from just numerical improvements, our results are interpretable and consistent with research in psychology showing that humans tend not to mention typical attributes \cite{MitchellEtAl13,WesterbeekEtAl15} unless required for unique identification~\cite{sedivy2003pragmatic} or distinguishability~\cite{olson1970language,tenenbaum1973locating}.

\section{Related work}
\begin{figure}[!t]
\centering
\includegraphics[width=0.45\textwidth]{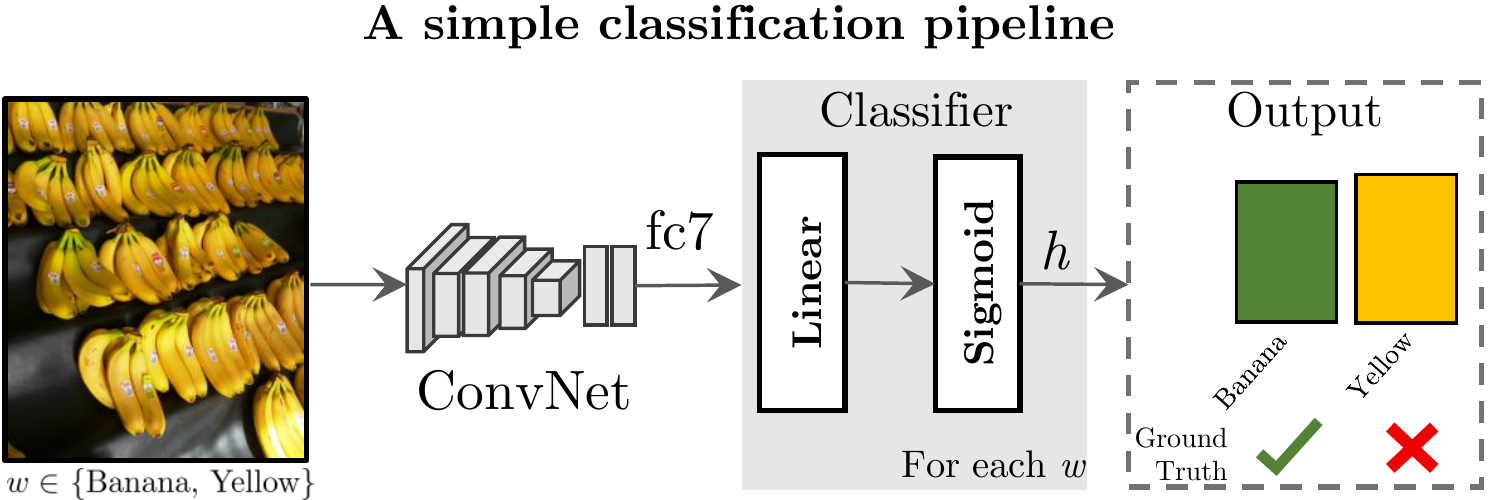}
\vspace{-0.08in}
\caption{A simple classification model for learning from human-centric annotations. The noisy labels (\texttt{banana} is not annotated as \texttt{yellow}) impede the learning process.}
\vspace{-0.1in}
\label{fig:simple-classification}
\end{figure}

\begin{figure*}[!t]
\includegraphics[width=1.0\textwidth]{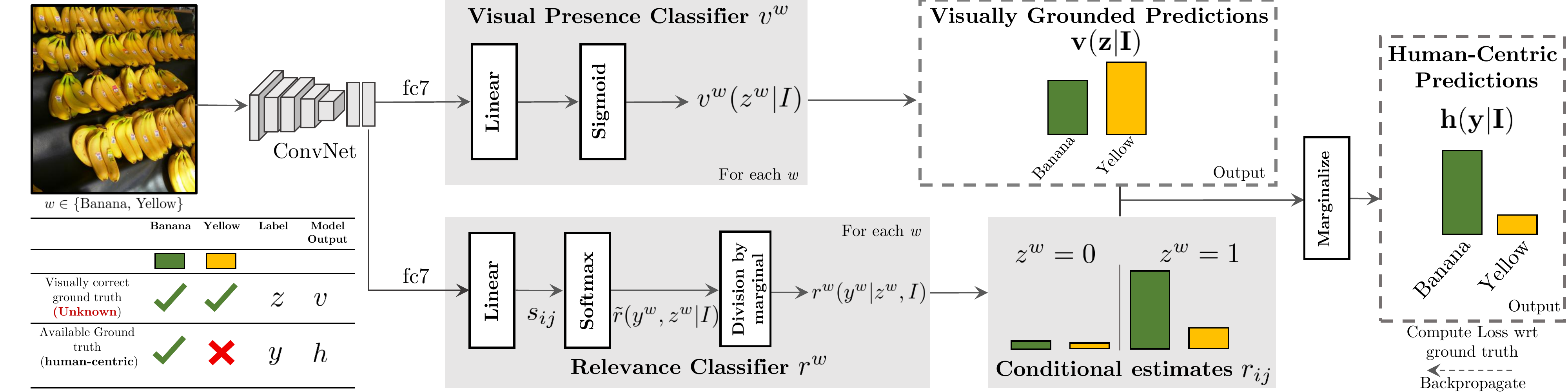}
\vspace{-0.08in}
\caption{Our model uses noisy human-centric annotations $y$ for learning visually grounded classifiers without access to the visually correct ground truth $z$. It uses two classifiers: a visual presence classifier $v$ and a relevance classifier $r$. The visual presence classifier $v$ predicts whether the visual concept $w$ is visually present in an image. The relevance classifier $r$ models the noise and predicts whether the concept should be mentioned or not. We combine these predictions to get the human-centric prediction $h$.}
\vspace{-0.08in}
\label{fig:alg-idea}
\end{figure*}

Label noise is ubiquitous in real world data. It can impact the training process of models and decrease their predictive accuracy~\cite{nettleton2010study,joulin2015learning,barnard2001learning}. Since there are vast amounts of cheaply available noisy data, learning good predictors despite the label noise is of great practical value.

The taxonomy of label noise presented in~\cite{frenay2014classification} differentiates between two broad categories of noise: \emph{noise at random} and \emph{statistically dependent noise}. The former does not depend on the data, while the latter does. In practice, one may encounter a combination of both types of noise.

Human-centric annotations~\cite{berg2012understanding} exhibit noise that is highly structured and shows statistical dependencies on the data~\cite{frenay2014classification,berg2012understanding,gazedescription}. It is structured in the sense that certain labels are preferentially omitted as opposed to others. Vision researchers have studied human-centric annotations in various settings, such as missing objects in image descriptions~\cite{berg2012understanding}, scenes~\cite{standsout}, and attributes~\cite{attributepops} and show that these annotations are noisy~\cite{berg2012understanding}.
Much of the work on learning from noisy labels focuses on robust algorithms~\cite{manwani2013noise, deeptagging}, voting methods~\cite{beigman2009learning}, or statistical queries~\cite{valiantnoise}. Some of these methods~\cite{deeptagging,valiantnoise} require access to clean oracle labels, which may not be readily available.

Explicitly modeling label noise has received increasing attention in recent years~\cite{noisy-labels, natrajan-noisy-labels, xiao2015learning, mnih2012learning}. Many of these methods operate under the ``noise at random'' assumption and treat noise as conditionally  independent of the image. \cite{larsen1998design} models symmetric label noise (independent of the true label), which is a strong assumption for real world data. \cite{natrajan-noisy-labels,noisy-labels} both model asymmetric label noise that is conditionally independent of the image. Such an assumption ignores the input image (and the objects therein) which directly affects the noisy annotations produced by humans~\cite{berg2012understanding}.

Recently, Xiao \etal~\cite{xiao2015learning} introduced an image conditional noise model that attempts to predict what type of noise corrupts each training sample (no noise, noise at random, and structured label swapping noise).
Unlike~\cite{xiao2015learning}, our training algorithm does not require a small amount of cleanly labeled training data to bootstrap parameter estimation. Our model is also specifically designed to handle the noise found in human-centric annotations.

Bootstrapping~\cite{reed2014training}, semi-supervised learning (SSL)~\cite{zhu2010semi,shrivastava2012constrained} and Positive Unlabeled (PU) learning~\cite{li2003learning,liu2003building,elkan2008learning,MisraSSL} are other ways of learning from noisy labeled data. However, they require access to clean oracle labels. SSL approaches are often computationally impractical~\cite{zhou2004learning,zhu2003semi} or make strong independence assumptions~\cite{fergus-ssl} that do not hold in human-centric annotations. Our approach, which trains directly on noisy labels, can serve as a starting point for these approaches.

The work described here is also consistent with research in psycholinguistics on object reference and description.  Such work demonstrates that humans store typical or ``prototypical'' representations of objects and their properties~\cite{rosch1999principles}; and this background knowledge appears to have an effect on object description~\cite{MitchellEtAl13,WesterbeekEtAl15}.  People tend not to mention attributes that are obvious or typical for an object, preferring to name attributes required for conversational relevance~\cite{Kronfeld89}, unique identification against alternatives~\cite{sedivy2003pragmatic,vanDerSluisEtAl06} and distinguishability~\cite{gregory1966,KoolenEtAl11}.  A similar separation between {\it what is observed} and {\it what is mentioned} falls out naturally from our proposed model.

\section{Our Approach}\label{sec:approach}
Our goal is to train visually grounded image classifiers for a set of visual concepts $w \in \mathcal{W}$ (\eg, \texttt{banana}, \texttt{yellow}, \texttt{zebra}) using images and their human-centric annotations.
The conventional approach to this problem, shown in Figure~\ref{fig:simple-classification}, is to naively apply a supervised learning algorithm:
train a classifier $h^w$ for each concept $w$, to predict its human-centric label $y^w \in \{0,1\}$ (not mentioned \vs mentioned) as a conditional probability distribution $h^w(y^w | \mathcal{I})$, in which $\mathcal{I}$ is an image.

The resulting classifier would attempt to mimic human reporting bias by predicting how a human would label the image $\mathcal{I}$ regardless of whether $w$ is visually present or absent. Thus, the predictions from each classifier $h^w$ will not be visually grounded and do not meet our goal. How can we build classifiers that predict whether a visual concept $w$ is present in an image and ``see through'' the noise in human-centric annotations?

\subsection{Factor decoupling}
\label{sec:factor-decoupling}
We propose to structure each concept output $h^w (y^w|\mathcal{I})$ in terms of two classifiers $v^w$ and $r^w$.
The first classifier $v^w$ models the conditional probability of the \emph{visual presence} of the concept $w$  in the image.
The second classifier $r^w$ models the conditional probability of the \emph{relevance} of the concept $w$, conditioned on the image and whether or not $w$ is estimated to be visually present.
The human-centric predictor is formed by marginalizing over the concept's visual presence, as described next.

Let $z^w \in \{0,1\}$ be a latent (or hidden) variable indicating whether the concept $w$ is visually present in an image.
Note that the training data only supplies human-centric labels $y^w$; the true values of $z^w$ are unknown during training.
For instance, in Figure~\ref{fig:teaser}{\color{red}(a)}, $y^w=1$ and $z^w=1$ when the bicycle is present and mentioned, while $y^w=0$ and $z^w=1$ in Figure~\ref{fig:teaser}{\color{red}(b)} when the bicycle is present but not mentioned.
We refer to $z^w$ as the \emph{visual presence label}.

The conditional probability of the human-centric label given an image $\mathcal{I}$, $h^w(y^w | \mathcal{I})$, can now be computed by marginalizing over the latent visual presence label $z^w$:

\begin{equation}
\label{eqn:decouple}
\hspace{-0.2em}h^w(y^w | \mathcal{I}) = \sum_{j \in \{ 0, 1 \}} r^w(y^w | z^w = j, \mathcal{I}) v^w(z^w = j| \mathcal{I}).
\end{equation}
An illustration of our model is shown in Figure~\ref{fig:alg-idea}. One important property of this formulation is that it allows the model to assign high confidence to unlabeled visual concepts: for an unmentioned concept ($y^w = 0$ and $z^w = 1$), the relevance classifier $r^w$ allows the visual presence classifier $v^w$ to assign a high probability to the true visual label ($z^w=1$) while still making a prediction that matches the human-centric label ($y^w=0$).
This property enables the model to ``see through'' human reporting bias.

To simplify notation, we drop the concept index $w$ from $y, z, h, v$ and $r$ when possible.
We denote the probability values of $r$ by:
\begin{align}
\label{eqn:noise-not-condition-image}
r_{ij} &= r(y=i|z=j, \mathcal{I}), ~~ \forall (i,j) \in \{0,1\}^2.
\end{align}

Another important property of the factorization in Equation \eqref{eqn:decouple} is that it provides a way to get two different predictions for the concept $w$ in the same image. The model can predict the visual presence of a visual concept; or predict how a human would annotate the image. Depending on the task at hand, one of these predictions may be more appropriate than the other, a point we demonstrate later via experimental results.

\subsection{Model learning and parameterization}
\label{sec:factor-decoupling-impl}

We estimate the model parameters by minimizing the regularized log loss
of $h^w$, summed over all concepts $w$, on the training data annotated with human-centric labels.
Since our model includes latent variables $z$ for each concept and image, one approach could be to use Expectation Maximization (EM) \cite{em}.
We choose a direct optimization approach~\cite{chapelle2014modeling} over EM for simplicity, and thus both the model parameters and latent variable posteriors are updated and inferred online. In each SGD minibatch, the model predicts the conditional distributions $r$ and $v$, marginalizes over the values of $z$, and uses the log loss of $h$ to drive better estimates of $r$ and $v$. 

The conditional distributions $r$ and $v$ are modeled with a ConvNet~\cite{fukushima1980neocognitron,rumelhart1985learning, lecun1989backpropagation}.
As illustrated in Figure~\ref{fig:alg-idea}, the ConvNet trunk is shared between the two distributions (per concept) and then branches near the output into two sets of untied parameters.
We jointly train one network for all visual concepts $w \in \mathcal{W}$ by treating learning as a multi-label classification problem.
Further network architecture details are given in Section \ref{sec:experiments} with experiments.

The conditional probability distribution $r$ models transition probabilities and thus its underlying joint distribution $\tilde{r}_{ij} = \tilde{r}(y = i, z = j | \mathcal{I})$ must be a valid probability distribution.
To enforce this constraint, we directly estimate the joint distribution $\tilde{r}$ with a softmax operation on a vector of unnormalized scores.

For each concept $w$, we first compute four scores $s_{ij}$ using four linear models parameterized by weights $m_{ij}$ and biases $b_{ij}$, and then normalize them using the softmax function to get a valid joint distribution $\tilde{r}_{ij}$:
\begin{align}
s_{ij} &= m_{ij}^{T}\phi(\mathcal{I}) + b_{ij}, \label{eqn:noise-linear} \\
\tilde{r}_{ij} &= \exp({s_{ij})} / \sum_{i'j'} \exp({s_{i'j'})}.
\end{align}
For $\phi(\mathcal{I})$, we use global image features computed by the shared ConvNet trunk (\eg, \texttt{fc7} layer activations from VGG16~\cite{VGG}).
These features capture the global image context, which is helpful in estimating $r$.
Each $r_{ij}$ can then be computed from $\tilde{r}$ by dividing by the marginal $\tilde{r}(z = j | \mathcal{I})$:
\begin{align}
r_{ij} &= \tilde{r}_{ij} / \sum_{i'}\tilde{r}_{i'j}.
\end{align}
Figure~\ref{fig:alg-idea} illustrates our process of computing $r_{ij}$ from the image.
Since our operations for estimating $\tilde{r}$ (and thus $r$) are differentiable, we can backpropagate their errors to the ConvNet, allowing the full model to be trained end-to-end.

\begin{table*}[!t]
\centering
\setlength{\tabcolsep}{0.36em}
\caption{mAP and PHR values on MS COCO captions ground truth (20k test images). We add our latent model to each baseline to make predictions that are visually grounded ($v$) or conform to human ($h$) labels. POS tags are as follows:  Nouns (NN), Verbs (VB), Adjectives (JJ), Determiners (DT), Pronouns (PRP), Prepositions (IN).}
\vspace{-0.08in}
\label{tbl:caption-gt-ap}
\resizebox{\linewidth}{!}{
\small {
\hspace*{-0.1in}
\begin{tabular}[b]{@{}cccccccccccccccccccc@{}}
\toprule
& & & \multicolumn{8}{c}{Mean Average Precision} & & \multicolumn{8}{c}{Precision at Human Recall} \\
\cmidrule{4-11}\cmidrule{13-20}
& & & NN & VB & JJ & DT & PRP & IN & Others & All & & NN & VB & JJ & DT & PRP & IN & Others & All \\
& &  Prob & 616 & 176 & 119 & 10 & 11 & 38 & 30 & 1000 & \multicolumn{2}{c}{\scriptsize{$\leftarrow$ Count}} & & & & & & \\
\midrule
\multirow{4}{*}{\rot[90]{VGG16}} & MILVC~\cite{mil-cvpr-15} & - & 41.6 & 20.7 & 23.9 & 33.4 & 20.4 & 22.5 & 16.3 & 34.0 & &
52.7 & 32.8 & 40.5 & 40.3 & 32.2 & 33.0 & 24.6 & 45.8 \\
& MILVC + Multiple-\texttt{fc8} & - & 41.1 & 20.9 & 23.7 & 33.6 & 21.1 & 22.8 & 16.8 & 33.8 & &
51.2 & 32.6 & 40.8 & 41.1 & 31.7 & 33.5 & 27.3 & 45.0 \\
& MILVC + Latent (Ours) & $v$ & 42.9 & 21.7 & 24.9 & 33.1 & 19.6 & 23.0 & 16.2 & 35.1 & &
53.6 & 35.4 & 43.3 & 41.3 & 28.0 & 36.0 & 24.4 & 47.2 \\
& MILVC + Latent (Ours) & $h$ & 44.3 & 22.3 & 25.8 & 34.4 & 21.8 & 23.6 & 17.3 & \textbf{36.3} & &
55.5 & 36.3 & 44.7 & 42.9 & 32.1 & 37.3 & 26.4 & \textbf{48.9} \\
\midrule
\multirow{3}{*}{\rot[90]{AlexNet}} & MILVC~\cite{mil-cvpr-15} & - & 33.2 & 16.2 & 20.1 & 30.9 & 16.4 & 19.9 & 14.6 & 27.4 & &
40.0 & 26.4 & 36.0 & 38.2 & 24.2 & 27.5 & 21.9 & 35.9 \\
& MILVC + Latent (Ours) & $v$ & 35.6 & 17.7 & 21.9 & 32.4 & 16.9 & 20.7 & 15.2 & 29.4 & &
43.9 & 28.3 & 37.5 & 41.2 & 29.2 & 29.9 & 23.3 & 39.0 \\
& MILVC + Latent (Ours) & $h$ & 36.5 & 18.0 & 22.4 & 32.9 & 17.8 & 21.4 & 15.6 & \textbf{30.1} & &
45.1 & 28.7 & 38.0 & 41.2 & 32.2 & 31.0 & 24.0 & \textbf{40.0} \\
\midrule
\multirow{4}{*}{\rot[90]{VGG16}} & Classif. & - & 34.9 & 18.1 & 20.5 & 32.8 & 19.2 & 21.8 & 16.3 & 29.0 & &
42.5 & 30.4 & 33.9 & 40.5 & 30.4 & 30.7 & 23.8 & 38.2 \\
& Classif. + Multiple-\texttt{fc8} & - & 34.2 & 17.7 & 19.9 & 32.6 & 19.0 & 21.5 & 15.9 & 28.4 & &
41.3 & 27.9 & 32.3 & 39.6 & 29.6 & 31.2 & 22.6 & 36.8 \\
& Classif. + Latent (Ours) & $v$ & 37.7 & 19.6 & 22.0 & 32.6 & 20.2 & 22.0 & 16.3 & 31.2 & &
46.3 & 32.9 & 36.8 & 38.9 & 32.3 & 33.1 & 27.0 & 41.5 \\
& Classif. + Latent (Ours) & $h$ & 38.7 & 20.1 & 22.6 & 33.8 & 21.2 & 23.0 & 17.5 & \textbf{32.0} & &
47.8 & 33.7 & 37.9 & 42.5 & 34.2 & 34.4 & 29.0 & \textbf{42.9} \\
\bottomrule
\end{tabular}
}
}
\end{table*}

\section{Experiments}
\label{sec:experiments}
\vspace{-0.08in}
\begin{figure}
\centering
\includegraphics[width=0.45\textwidth]{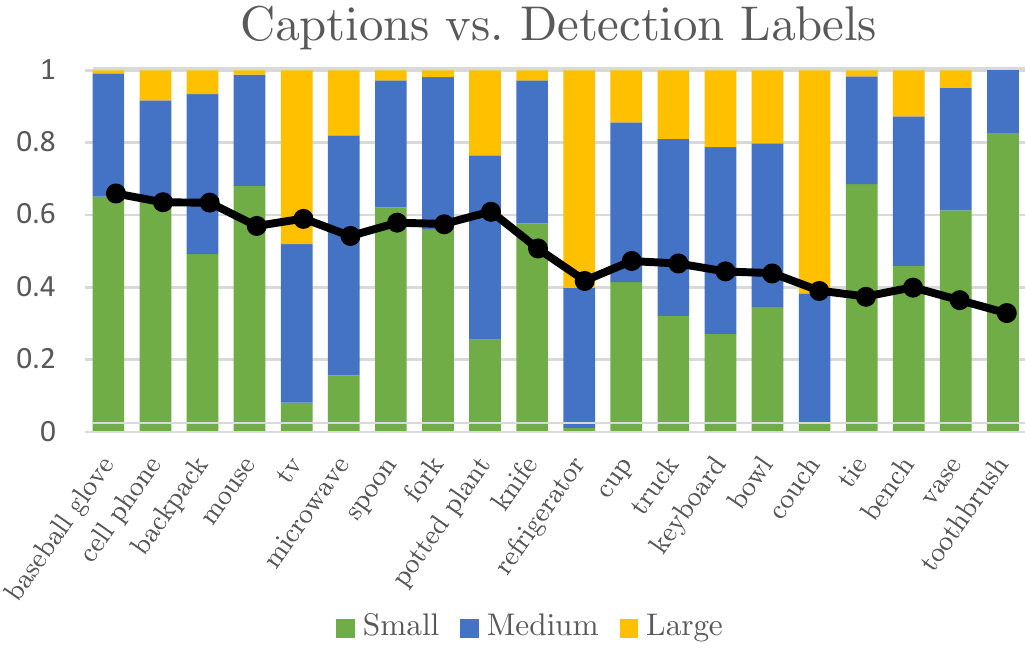}
\vspace{-0.08in}
\caption{We use the MS COCO dataset to display the objects with the highest reporting bias. We use the detection labels for objects and see whether they are mentioned in the caption. The black line shows the probability of an object not being mentioned in the caption. The distribution of sizes for objects that are missed are shown by the color bars. Note that for many categories most unreported objects are small or medium in size (green and blue).}
\vspace{-0.08in}
\label{fig:annot-mismatch}
\end{figure}

We evaluate our proposed model on two datasets: Microsoft COCO~\cite{COCO} and a random subset of the Yahoo Flickr Creative Commons 100M (YFCC100M) dataset~\cite{yfcc100m}.
The YFCC100M dataset includes user-generated image tags, which we take as our source of human-centric annotations.
For MS COCO, we take the supplied image captions~\cite{COCOCaption} as human-centric annotations.
We use the MS COCO object detection labels for dataset analysis and algorithm evaluation.
These labels allow us to verify the accuracy of trained visual presence classifiers, $v^w$; these labels are never used for training our model.

\subsection{Experiments on MS COCO 1k visual concepts}
Our first set of experiments use the MS COCO 1000 visual concepts from \cite{mil-cvpr-15}.
The visual concepts are the 1000 most common words in the MS COCO captions dataset~\cite{COCOCaption} and include nouns, verbs, adjectives, and other parts of speech (see Table \ref{tbl:caption-gt-ap} for a breakdown).

For training, we generate image labels as 1000-dimensional binary vectors indicating which of the 1000 target visual concepts are present in the any of the 5 reference captions for each training image.
The training set includes approximately 80k images.
For evaluation, we follow~\cite{mil-cvpr-15} and split the val set into equally sized val and test sets of $\sim$20k images each; we use the same splits as in~\cite{mil-cvpr-15}. We report results on this 20k image test set.

\subsubsection{Human reporting bias in image descriptions}
We first analyze the annotation mismatch between the caption labels and the detection labels. We obtain labels for the 73 objects common in both the caption and object detection labels (see Section~\ref{sec:visual-predictions} for details). We use the notation from Section~\ref{sec:factor-decoupling}, and measure the human reporting bias as $r^{\star}_{01}$: the probability of an object not being mentioned in the caption ground truth ($y=0$) and being present in the detection ground truth ($z=1$), over all the training images. To account for object size as a factor in an object not being mentioned, we split these measurements based on the size of the bounding box (sizes as defined in~\cite{COCO}). Figure~\ref{fig:annot-mismatch} shows this mismatch for the top 20 objects with the highest $r^{\star}_{01}$ for values. A high $r^{\star}_{01}$ value indicates that there is a large mismatch between the caption ground truth and the detection ground truth -- objects that are visually present but not mentioned in the captions.

As observed, there is a high degree of human labeling noise in the image descriptions, with the object with highest reporting bias mentioned roughly half as much as it appears.

\subsubsection{Evaluating human-centric label prediction}
\label{sec:coco-caption}
We can evaluate our model in two ways: as a purely visual classifier ($v$) or as a predictor of human-centric ($h$) labels. We start with the latter and evaluate our model's predictions against the human-centric MS COCO captions.

As a strong baseline, we use the recently proposed MILVC~\cite{mil-cvpr-15} approach. This method applies a ConvNet (VGG16~\cite{VGG} or AlexNet \cite{alexnet})\footnote{Unless otherwise specified, we use VGG16 for all our experiments.} in a fully-convolutional way to a large input image to generate a $12 \times 12$ grid of human-centric label predictions. It then uses a noisy-OR \cite{Viola06multipleinstance} to compute a single prediction from the 144 intermediate values. During training, the noisy-OR induces a form of multiple instance learning (MIL). Note that the baseline model estimates $h$ labels directly without our decomposition into relevance and visual presence factors. As a second baseline (Classif.), we use a vanilla classification model akin to Figure~\ref{fig:simple-classification}, in which an ImageNet~\cite{ILSVRC15} pre-trained ConvNet is fine-tuned to directly predict human-centric labels.

We also include an additional baseline variant that adds extra parameters to control for the fact that our proposed model requires adding extra parameters. Specifically, we train a ``Multiple-\texttt{fc8}'' model for each method (MILVC and Classif.)~that has the same number of parameters as our model. To train Multiple-\texttt{fc8}, we add four extra randomly initialized \texttt{fc8} (linear classification) layers, each with their own loss. At test time, we average the predictions of all the \texttt{fc8} layers to get the final prediction.

We implement two variants of our model to parallel the MILVC and Classif.~baselines.
In the first variant (MILVC + Latent), $v$ uses a noisy-OR over a $12 \times 12$ grid of visual presence predictions, and for $r$ we average pool the $144$ \texttt{fc7} activation vectors to obtain a single 4096 dimensional $\phi(\mathcal{I})$ for Equation~\ref{eqn:noise-linear}.
The second variant (Classif.~+ Latent) generates a $1 \times 1$ output for $v$ and \texttt{fc7}, and therefore omits the noisy-OR and average pooling.
In both cases, $h$ predictions are obtained following Equation~\ref{eqn:decouple}, using both $r$ and $v$.
Like the baselines, our model is trained to minimize the log loss (cross-entropy loss) over $h$.

To train our model, we set the joint noise distribution $\tilde{r}$ to identity (\ie, $\tilde{r}_{11} = \tilde{r}_{00} = 0.5$) for the first two epochs and then update it for the last two epochs. Table~\ref{tbl:caption-gt-ap} shows mean average precision (mAP)~\cite{PASCAL} and precision at human recall (PHR)~\cite{COCOCaption} on the 20k test set. PHR is a metric proposed in~\cite{COCOCaption}, and measures precision based on human agreement. Briefly, this metric uses multiple references per image to compute a ``human recall'' value, an estimate of the probability that a human will use a particular word for an image.  Precision is then computed at this ``human recall'' value to get PHR.~\cite{COCOCaption} shows that for the task of predicting visual concepts, PHR is a more stable metric than an AP metric, since it accounts for human agreement.

We report results for the 1000 visual concepts in aggregate, as well as grouped by their part-of-speech (POS) tags, on the MS COCO test split of 20k images (Table~\ref{tbl:caption-gt-ap}). Our latent variable model improves classification performance over all the baseline networks and architectures by \textbf{3 to 4 points} for both metrics (mAP and PHR). Interestingly, the Multiple-\texttt{fc8} model, which has the same number of parameters as our latent model, does not show an improvement, even after extensive tuning of learning hyperparameters.
This finding makes the contribution of the proposed model evident; the improvement is not simply due to adding extra parameters.
It is worth noting that in Table \ref{tbl:caption-gt-ap}, $h$ is a better predictor of MS COCO caption labels than $v$, as $h$ directly models the human-centric labels used for evaluation.

\subsubsection{Evaluating visual presence prediction}
\label{sec:visual-predictions}
The decoupling of visual presence ($v$) predictions and human-centric ($h$) label predictions allows our model to learn better visual predictors. To demonstrate this, we use the fully-labeled ground truth from the COCO detection annotations to evaluate the visually grounded $v$ label predictions.

Since the 1000 visual concepts include many fine-grained visual categories (\eg, {\prop man}, {\prop woman}, {\prop child}) and synonyms (\eg, {\prop bike}, {\prop bicycle}), 
we manually specify a mapping from the visual concepts to the 80 MS COCO detection categories, \eg, $\{${\prop bike}, {\prop bicycle}$\}$ $\rightarrow$ {\prop bicycle}. We find that 73 of the 80 detection categories are present in the 1000 visual concepts. We use this mapping only at evaluation time to compute the probability of a detection category as the maximum of the probabilities of its fine-grained/synonymous categories.

Table~\ref{tbl:coco-detection-results} shows the mean average precision (mAP) of our method, as well as the baseline on these 73 categories.
As expected, using the human-centric model ($h$) for this task of visual prediction hurts performance (slightly).
The performance drop is less dramatic when evaluating on these 73 classes because these classes have less label noise for large sized objects as compared to the 1000 visual concepts.
We also train a noise-free reference model using the ground-truth visual labels from the detection dataset (\ie, the true values of the latent $z$ labels).
\begin{table}[!t]
\centering
\setlength{\tabcolsep}{0.42em}
\vspace{-0.08in}
\caption{Visual classification on 73 classes evaluated using fully labeled data from COCO.}
\vspace{-0.08in}
\label{tbl:coco-detection-results}
\begin{tabular}[b]{@{}ccccc@{}}
\toprule
& MILVC~\cite{mil-cvpr-15} & $v$ & $h$ & Using ground truth\\
\midrule
mAP & 63.7 & \textbf{66.8} & 66.5 & 76.3\\
\bottomrule
\end{tabular}
\end{table}

\subsubsection{Importance of conditioning on input images}
A central point of this paper (and also in \cite{berg2012understanding}) is that human-centric label noise is statistically dependent on image data.
Here we demonstrate that our model is indeed improved by conditioning the noise (relevance) distribution on the input image, in contrast to previous work~\cite{natrajan-noisy-labels,noisy-labels} that estimates noise parameters \emph{without} conditioning on the image.

To better understand the importance of this conditioning, we consider a model akin to~\cite{noisy-labels}. We estimate the latent distribution $r$ without conditioning on the input image, and compare it to our model that computes $r$ conditioned on the image. Table~\ref{tbl:no-image} shows that mAP is significantly improved by conditioning on the image. When not conditioned on the image, only minor gains are achieved.

\begin{table}[!t]
\centering
\setlength{\tabcolsep}{0.42em}
\vspace{-0.08in}
\caption{We show the importance of conditioning the relevance $r$ on the input image. We measure the classification mAP on MS COCO 1k visual concepts using both the visually grounded ($v$) and the human-centric ($h$) predictions. Conditioning on the input image shows improvement over the baseline showing that human reporting bias statistically depends on the input image.}
\vspace{-0.08in}
\label{tbl:no-image}
\begin{tabular}[b]{@{}ccccccc@{}}
\toprule
& & \multicolumn{2}{c}{w/o image} & & \multicolumn{2}{c}{w/ image} \\
\cmidrule{3-4}\cmidrule{6-7}
& MILVC~\cite{mil-cvpr-15}& $v$ & $h$ & &  $v$ & $h$ \\
\midrule
mAP & 34.0 & 34.2 & 34.3 & &  35.1 & \textbf{36.3} \\
\bottomrule
\end{tabular}
\vspace{-0.08in}
\end{table}

\begin{figure*}[!t]
\centering
\includegraphics[width=0.98\linewidth]{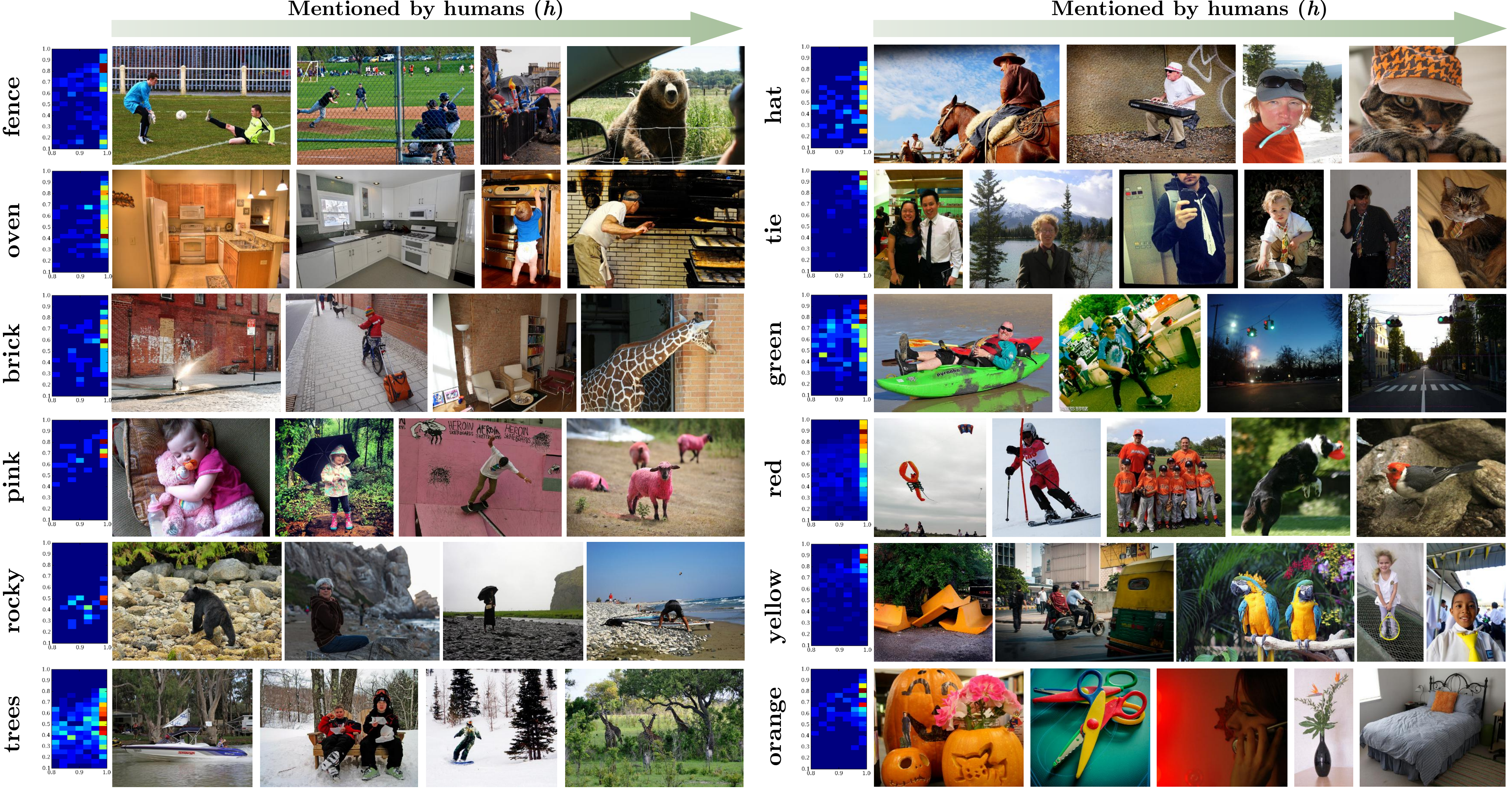}
\vspace{-0.11in}
\caption{Our model modifies visually correct detections to conform to human labeling. We show this modification for a few images of target visual concepts in the MS COCO Captions dataset. We first show the variation between $h$  (y axis) and $v$ values (x axis) for each concept in a 2D histogram. After thresholding at $v \geq 0.8$, we pick a representative image from each quantile of $h$ ($h$ increases from left to right).
As you move from left to right, the model transitions from predicting that a human would not ``speak'' the word to predicting that a human would speak it.
The human-centric $h$ predictions of concepts depend on the image context, \eg, {\prop fence} at a soccer game vs. {\prop fence} between a bear and a human (first row). Our model picks up such signals to not only learn a visually correct {\prop fence} predictor, but also when a {\prop fence} should be mentioned.}
\vspace{-0.13in}
\label{fig:results-vocab}
\end{figure*}

\begin{figure*}[!t]
\centering
\includegraphics[width=0.98\linewidth]{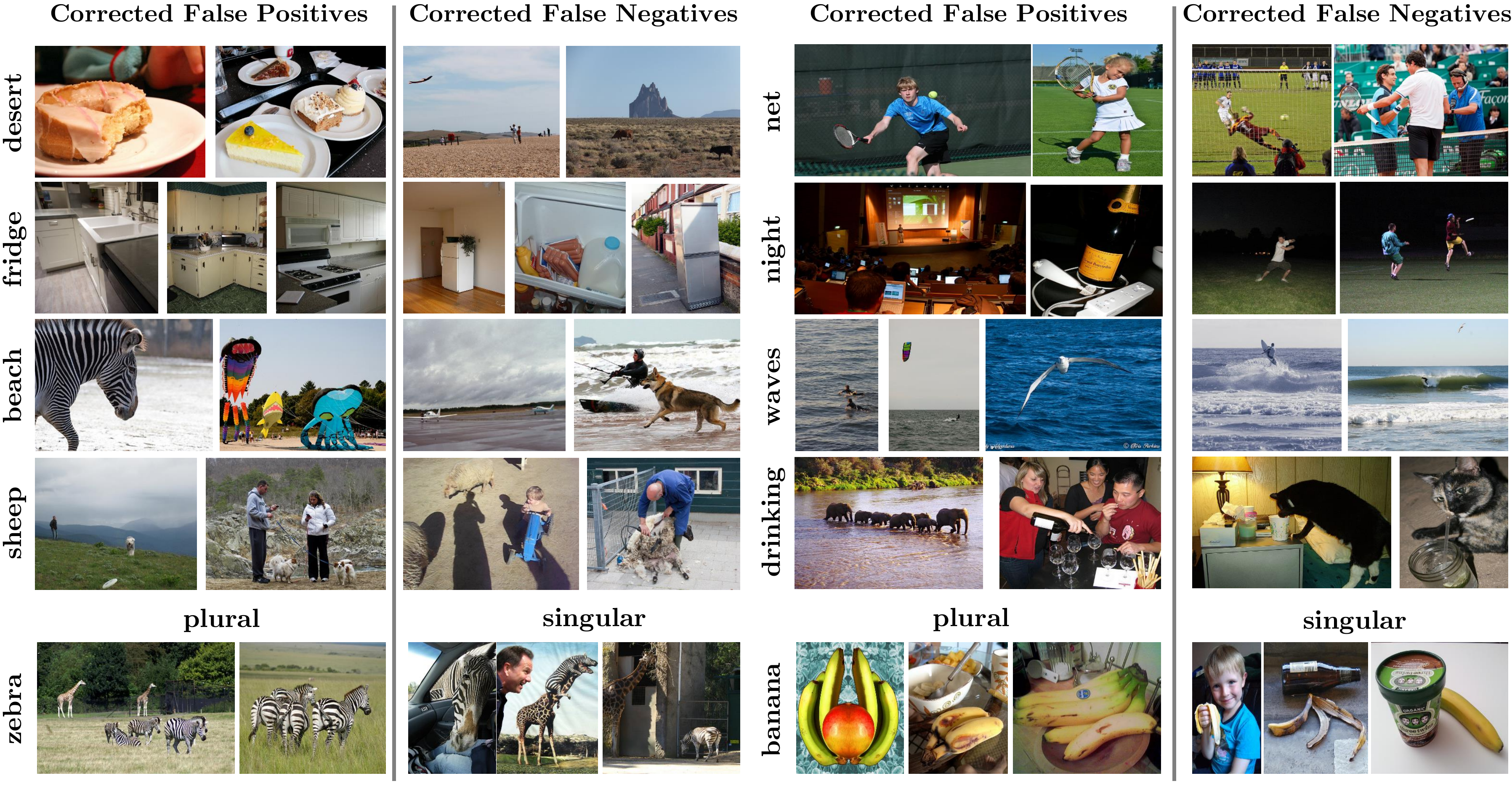}
\vspace{-0.11in}
\caption{Our model learns clean visual predictors from noisy labels.
Here we show corrected false positives: MILVC incorrectly reports a high probability ($h \ge 0.75$) for the concept, while our model correctly reports a low probability ($v \le 0.3$); and corrected false negatives: MILVC incorrectly reports a low probability ($h \le 0.3$) for the concept, while our model correctly reports a high probability ($v \ge 0.75)$. For example, consider \texttt{zebra} vs. \texttt{zebras}, and \texttt{banana} vs. \texttt{bananas} in the last row, where our model correctly ``counts'' compared to the baseline.
Images are from the MS COCO Captions dataset.}
\vspace{-0.13in}
\label{fig:results-correction}
\end{figure*}

\begin{table}[!b]
\centering
\vspace{-0.08in}
\caption{mAP values on a subset of YFCC100M. We add our latent model over the MILVC baseline to make predictions that are visually grounded ($v$) or that conform to human-centric ($h$) labels. POS tags: Nouns (NN), Verbs (VB), Adjectives (JJ), Pronouns (PRP), Prepositions (IN).}
\vspace{-0.08in}
\label{tbl:flickr-tag}
\resizebox{\linewidth}{!}{
\setlength{\tabcolsep}{0.3em}
\small
\begin{tabular}[b]{@{}cccccccccc@{}}
\toprule
& & \multicolumn{7}{c}{\hspace{0.55in}Mean Average Precision} \\
\cmidrule{4-10}
&& & NN & VB & JJ  & PRP & IN & Others & All  \\
& \scriptsize{Count $\rightarrow$} & Prob &  791 & 10 & 148  & 13 & 23 & 15 & 1000\\
\midrule
\multirow{4}{*}{\rot[90]{VGG16}} & MILVC~\cite{mil-cvpr-15} & - &  5.7 & 9.2 & 5.2 & 3.8 & 8.8 & 6.1 & 5.7 \\
& MILVC + Multiple-\texttt{fc8} & - & 4.6 & 6.2 & 3.8 & 2.7 & 7.3 & 3.1 & 4.5 \\
& MILVC + Latent (Ours) & $v$ & 9.8 & 15.1 & 8.9 & 8.3 & 12.4 & 12.4 & 9.8 \\
& MILVC + Latent (Ours) & $h$ & 11.2 & 15.4 & 9.9 & 8.2 & 16.3 & 12.5 & \textbf{11.2} \\
\bottomrule
\end{tabular}
}
\end{table}

\subsection{Experiments on Flickr image tagging}
Datasets like MS COCO are curated by searching for images with specific objects~\cite{COCO}.
In contrast, social media websites like Flickr contain much larger collections of images that are annotated with user-generated content such as tags, keywords, and descriptions.
The words found in such data exhibit the human-centric annotation properties modeled by our approach.

We test our model on this ``real world'' data by using a random subset of $\sim$89k images from the YFCC100M dataset~\cite{yfcc100m}.
We ensure that these images have at least 5 and at most 30 human annotated tags that are present in the WordNet~\cite{wordnet} lexicon. We split this dataset into 75k training images and 14k test images, and consider the top 1000 tags as the set of visual concepts. We train the baseline MILVC~\cite{mil-cvpr-15} model and our model for 4 epochs following the same hyperparameters used for MS COCO training.

Table~\ref{tbl:flickr-tag} shows the numerical results of these models evaluated on the test set using the same human annotated tags. As explained in Section~\ref{sec:coco-caption}, we compare against the MILVC baseline, and a model with the same number of parameters as ours (denoted by Multiple-\texttt{fc8}). Our model has double the performance of the baseline MILVC model and \textbf{increases mAP by 5.5 points}.


\begin{table}[!b]
\centering
\caption{LSTM captioning results on MS COCO}
\vspace{-0.08in}
\label{tbl:captioning}
\footnotesize{
\begin{tabular}[b]{@{}ccccc@{}}
\toprule
& Prob & BLEU-4 & ROUGE & CIDEr \\
\midrule
MILVC~\cite{mil-cvpr-15} & - & 27.7 & 51.8 & 89.7\\
MILVC + Latent (Ours) & $h$ & \textbf{29.2} & \textbf{52.4} & \textbf{92.8} \\
\bottomrule
\end{tabular}
}
\end{table}

\subsection{Interpretability of the noise model}
\vspace{-0.08in}
The relevance classifier $r$ models human labeling noise conditioned on the image.
Depending on the image, it can enhance or suppress the visual prediction for each concept.
We show such modifications for a few visual concepts in Figure~\ref{fig:results-vocab}.
After thresholding at $v \geq 0.8$, we pick a representative image from each quantile of $h$ ($h$ increases from left to right). 
The variation in $h$ values for these high confidence $v \geq 0.8$ images (shown in a 2D histogram in each row) indicates that $h$ and $v$ have been decoupled by our model.
The images show that our model captures subtle nuances in the ground truth, \eg, mention a hat worn by a cat, do not mention the color of a pumpkin, definitely mention pink sheep, \etc.
It automatically captures that context is important for certain objects like {\prop fence} and {\prop hat}, while certain attributes are worth mentioning to help distinguish objects like the {\prop orange} pillow.
Such connections have been shown in both vision research~\cite{berg2012understanding} and psychology~\cite{sedivy2003pragmatic, gregory1966}.

\subsection{Correcting error modes by decoupling}
Modeling latent noise in human-centric annotations allows us to learn clean visual classifiers. In Figure~\ref{fig:results-correction}, we compare our model's visual presence $v$ predictions with the baseline (MILVC) and show a few error modes that it corrects. Our model is able to correct error modes like misspellings ({\prop desert} vs.~{\prop dessert} in the first row), localizes objects correctly and out of context ({\prop fridge} in the second row, {\prop net} in the first row, \etc) and is better at counting ({\prop zebra}, {\prop banana} last row).

\subsection{Using word detections for caption generation}
\vspace{-0.08in}
We now look at the task of automatic image caption generation and show how our model can help improve the task. We consider a basic Long Short-Term Memory (LSTM)~\cite{lstm} network to generate captions. We use 1000 cells for the LSTM, and learn a 256 dimensional word embedding for the input words. Following~\cite{lrcn}, our vocabulary consists of words with frequency $\geq 5$ in the input captions. The image features (1000 visual concept probabilities) are fed once to the LSTM as its first hidden input. We train this LSTM over all the captions in the MS COCO caption training data for 20 epochs using~\cite{nlpcaffe,caffe}. We use beam size of 1 for decoding. Table~\ref{tbl:captioning} shows the evaluation of the automatically generated captions using standard captioning metrics. Using the probabilities from our model shows an improvement for all evaluation metrics. Thus, modeling the human-reporting bias can help downstream applications that require such human-centric predictions.

\section{Discussion}

We have introduced an algorithm that explicitly models {\it reporting bias} --- the discrepancy between what exists and what people mention --- for image labeling.  By introducing a latent variable to capture ``what is in an image'' separate from ``what is labeled in an image'', we leverage {\it human-centric} annotations of images to their full potential, inferring visual concepts present in an image separately from the visual concepts worth mentioning. We demonstrate performance improvements over previous work on several tasks, including image classification and image captioning. Further, the proposed model is highly interpretable, capturing which concepts may be included or excluded based on the context and dependencies across visual concepts. Initial inspection of the model's predictions suggests consistency with psycholinguistic research on object description, with typical properties noticed but not mentioned.

The algorithm and techniques discussed here pave the way for new deep learning methods that decouple human performance from algorithmic understanding, modeling both jointly in a network that can be trained end-to-end.  Future work may explore different methods to incorporate constraints on the latent variables, or to estimate their posteriors (such as with EM). Finally, to fully exploit the enormous amounts of data which exist ``in the wild'', algorithms that explicitly handle noisy data are essential.

{\small
\par \noindent \textbf{Acknowledgments:}
We thank Jacob Devlin, Lucy Vanderwende, Frank Ferraro, Sean Bell, Abhinav Shrivastava, and Saurabh Gupta for helpful discussions. Devi Parikh and Dhruv Batra for their suggestions and organizing the fun `snack times'.
}

{\small
\bibliographystyle{ieee}
\bibliography{myRefs}

\begin{thebibliography}{10}\itemsep=-1pt

\bibitem{barnard2001learning}
K.~Barnard and D.~Forsyth.
\newblock Learning the semantics of words and pictures.
\newblock In {\em ICCV}, 2001.

\bibitem{beigman2009learning}
E.~Beigman and B.~B. Klebanov.
\newblock Learning with annotation noise.
\newblock In {\em ACL-IJCNLP}, 2009.

\bibitem{berg2012understanding}
A.~C. Berg, T.~L. Berg, H.~Daume~III, J.~Dodge, A.~Goyal, X.~Han, A.~Mensch,
  M.~Mitchell, A.~Sood, K.~Stratos, et~al.
\newblock Understanding and predicting importance in images.
\newblock In {\em CVPR}, 2012.

\bibitem{standsout}
A.~Borji, D.~N. Sihite, and L.~Itti.
\newblock What stands out in a scene? a study of human explicit saliency
  judgment.
\newblock {\em Vision research}, 91, 2013.

\bibitem{chapelle2014modeling}
O.~Chapelle.
\newblock Modeling delayed feedback in display advertising.
\newblock In {\em KDD}, 2014.

\bibitem{COCOCaption}
X.~Chen, H.~Fang, T.-Y. Lin, R.~Vedantam, S.~Gupta, P.~Dollar, and C.~L.
  Zitnick.
\newblock Microsoft coco captions: Data collection and evaluation server.
\newblock {\em arXiv preprint arXiv:1504.00325}, 2015.

\bibitem{em}
A.~P. Dempster, N.~M. Laird, and D.~B. Rubin.
\newblock Maximum likelihood from incomplete data via the em algorithm.
\newblock {\em Journal of the royal statistical society. Series B
  (methodological)}, pages 1--38, 1977.

\bibitem{ImageNet}
J.~Deng, W.~Dong, R.~Socher, L.~jia Li, K.~Li, and L.~Fei-fei.
\newblock Imagenet: A large-scale hierarchical image database.
\newblock In {\em CVPR}, 2009.

\bibitem{lrcn}
J.~Donahue, L.~A. Hendricks, S.~Guadarrama, M.~Rohrbach, S.~Venugopalan,
  K.~Saenko, and T.~Darrell.
\newblock Long-term recurrent convolutional networks for visual recognition and
  description.
\newblock In {\em CVPR}, 2015.

\bibitem{elkan2008learning}
C.~Elkan and K.~Noto.
\newblock Learning classifiers from only positive and unlabeled data.
\newblock In {\em SIGKDD}, 2008.

\bibitem{PASCAL}
M.~Everingham, L.~Van~Gool, C.~K.~I. Williams, J.~Winn, and A.~Zisserman.
\newblock The {PASCAL} {V}isual {O}bject {C}lasses {C}hallenge 2007
  {(VOC2007)}.

\bibitem{mil-cvpr-15}
H.~Fang, S.~Gupta, F.~N. Iandola, R.~Srivastava, L.~Deng, P.~Doll{\'{a}}r,
  J.~Gao, X.~He, M.~Mitchell, J.~C. Platt, C.~L. Zitnick, and G.~Zweig.
\newblock From captions to visual concepts and back.
\newblock In {\em CVPR}, 2015.

\bibitem{fergus-ssl}
R.~Fergus, Y.~Weiss, and A.~Torralba.
\newblock Semi-supervised learning in gigantic image collections.
\newblock In {\em NIPS}, 2009.

\bibitem{frenay2014classification}
B.~Fr{\'e}nay and M.~Verleysen.
\newblock Classification in the presence of label noise: a survey.
\newblock {\em NNLS}, 25, 2014.

\bibitem{fukushima1980neocognitron}
K.~Fukushima.
\newblock Neocognitron: A self-organizing neural network model for a mechanism
  of pattern recognition unaffected by shift in position.
\newblock {\em Biological cybernetics}, 36(4):193--202, 1980.

\bibitem{gordon2013reportingBias}
J.~Gordon and B.~V. Durme.
\newblock Reporting bias and knowledge extraction.
\newblock In {\em Automated Knowledge Base Construction (AKBC) 2013: The 3rd
  Workshop on Knowledge Extraction, at CIKM 2013}, AKBC'13, 2013.

\bibitem{gregory1966}
R.~L. Gregory.
\newblock Eye and brain: The psychology of seeing.
\newblock 1966.

\bibitem{lstm}
S.~Hochreiter and J.~Schmidhuber.
\newblock Long short-term memory.
\newblock {\em Neural computation}, 9(8):1735--1780, 1997.

\bibitem{deeptagging}
H.~Izadinia, B.~C. Russell, A.~Farhadi, M.~D. Hoffman, and A.~Hertzmann.
\newblock Deep classifiers from image tags in the wild.
\newblock In {\em Workshop on Community-Organized Multimodal Mining:
  Opportunities for Novel Solutions}. ACM, 2015.

\bibitem{caffe}
Y.~Jia, E.~Shelhamer, J.~Donahue, S.~Karayev, J.~Long, R.~Girshick,
  S.~Guadarrama, and T.~Darrell.
\newblock Caffe: Convolutional architecture for fast feature embedding.
\newblock In {\em ACMM}, 2014.

\bibitem{joulin2015learning}
A.~Joulin, L.~van~der Maaten, A.~Jabri, and N.~Vasilache.
\newblock Learning visual features from large weakly supervised data.
\newblock {\em arXiv preprint arXiv:1511.02251}, 2015.

\bibitem{valiantnoise}
M.~Kearns.
\newblock Efficient noise-tolerant learning from statistical queries.
\newblock {\em JACM}, 45, 1998.

\bibitem{KoolenEtAl11}
R.~Koolen, A.~Gatt, M.~Goudbeek, and E.~Krahmer.
\newblock {\em Journal of Pragmatics}, 43(13):3231--3250, 2011.

\bibitem{alexnet}
A.~Krizhevsky, I.~Sutskever, and G.~E. Hinton.
\newblock Imagenet classification with deep convolutional neural networks.
\newblock In {\em NIPS}, 2012.

\bibitem{Kronfeld89}
A.~Kronfeld.
\newblock Conversationally relevant descriptions.
\newblock {\em Proceedings of the 27th Annual Meeting of the Association for
  Computational Linguistics}, 1989.

\bibitem{larsen1998design}
J.~Larsen, L.~Nonboe, M.~Hintz-Madsen, and L.~K. Hansen.
\newblock Design of robust neural network classifiers.
\newblock In {\em Acoustics, Speech and Signal Processing}, volume~2, 1998.

\bibitem{lecun1989backpropagation}
Y.~LeCun, B.~Boser, J.~S. Denker, D.~Henderson, R.~E. Howard, W.~Hubbard, and
  L.~D. Jackel.
\newblock Backpropagation applied to handwritten zip code recognition.
\newblock {\em Neural computation}, 1(4):541--551, 1989.

\bibitem{li2003learning}
X.~Li and B.~Liu.
\newblock Learning to classify texts using positive and unlabeled data.
\newblock In {\em IJCAI}, volume~3, 2003.

\bibitem{COCO}
T.-Y. Lin, M.~Maire, S.~Belongie, J.~Hays, P.~Perona, D.~Ramanan,
  P.~Doll{\'a}r, and C.~L. Zitnick.
\newblock Microsoft coco: Common objects in context.
\newblock In {\em ECCV}. 2014.

\bibitem{liu2003building}
B.~Liu, Y.~Dai, X.~Li, W.~S. Lee, and P.~S. Yu.
\newblock Building text classifiers using positive and unlabeled examples.
\newblock In {\em ICDM}, 2003.

\bibitem{MitchellEtAl13}
M.~M., R.~E., and van Deemter~K.
\newblock Typicality and object reference.
\newblock 2013.

\bibitem{manwani2013noise}
N.~Manwani and P.~Sastry.
\newblock Noise tolerance under risk minimization.
\newblock {\em Cybernetics}, 43, 2013.

\bibitem{wordnet}
G.~A. Miller.
\newblock Wordnet: a lexical database for english.
\newblock {\em Communications of the ACM}, 38(11):39--41, 1995.

\bibitem{MisraSSL}
I.~Misra, A.~Shrivastava, and M.~Hebert.
\newblock Watch and learn: Semi-supervised learning of object detectors from
  videos.
\newblock In {\em CVPR}, 2015.

\bibitem{mnih2012learning}
V.~Mnih and G.~E. Hinton.
\newblock Learning to label aerial images from noisy data.
\newblock In {\em ICML}, 2012.

\bibitem{natrajan-noisy-labels}
N.~Natarajan, I.~S. Dhillon, P.~K. Ravikumar, and A.~Tewari.
\newblock Learning with noisy labels.
\newblock In {\em NIPS}. 2013.

\bibitem{nettleton2010study}
D.~F. Nettleton, A.~Orriols-Puig, and A.~Fornells.
\newblock A study of the effect of different types of noise on the precision of
  supervised learning techniques.
\newblock {\em Artificial intelligence review}, 33, 2010.

\bibitem{nlpcaffe}
{NLP}caffe.
\newblock \url{http://github.com/Russell91/NLPCaffe}.

\bibitem{olson1970language}
D.~R. Olson.
\newblock Language and thought: Aspects of a cognitive theory of semantics.
\newblock {\em Psychological review}, 77, 1970.

\bibitem{reed2014training}
S.~Reed, H.~Lee, D.~Anguelov, C.~Szegedy, D.~Erhan, and A.~Rabinovich.
\newblock Training deep neural networks on noisy labels with bootstrapping.
\newblock {\em arXiv preprint arXiv:1412.6596}, 2014.

\bibitem{rosch1999principles}
E.~Rosch.
\newblock Principles of categorization.
\newblock {\em Concepts: core readings}, pages 189--206, 1999.

\bibitem{rumelhart1985learning}
D.~E. Rumelhart, G.~E. Hinton, and R.~J. Williams.
\newblock Learning internal representations by error propagation.
\newblock Technical report, DTIC Document, 1985.

\bibitem{ILSVRC15}
O.~Russakovsky, J.~Deng, H.~Su, J.~Krause, S.~Satheesh, S.~Ma, Z.~Huang,
  A.~Karpathy, A.~Khosla, M.~Bernstein, A.~C. Berg, and L.~Fei-Fei.
\newblock {ImageNet Large Scale Visual Recognition Challenge}.
\newblock {\em IJCV}, 115, 2015.

\bibitem{sedivy2003pragmatic}
J.~C. Sedivy.
\newblock Pragmatic versus form-based accounts of referential contrast:
  Evidence for effects of informativity expectations.
\newblock {\em Journal of psycholinguistic research}, 32(1), 2003.

\bibitem{shrivastava2012constrained}
A.~Shrivastava, S.~Singh, and A.~Gupta.
\newblock Constrained semi-supervised learning using attributes and comparative
  attributes.
\newblock In {\em ECCV}. 2012.

\bibitem{VGG}
K.~Simonyan and A.~Zisserman.
\newblock Very deep convolutional networks for large-scale image recognition.
\newblock {\em arXiv preprint arXiv:1409.1556}, 2014.

\bibitem{noisy-labels}
S.~Sukhbaatar, J.~Bruna, M.~Paluri, L.~Bourdev, and R.~Fergus.
\newblock Training convolutional networks with noisy labels.
\newblock In {\em ICLR Workshop}, 2015.

\bibitem{tenenbaum1973locating}
J.~M. Tenenbaum.
\newblock On locating objects by their distinguishing features in multisensory
  images.
\newblock {\em Computer Graphics and Image Processing}, 2, 1973.

\bibitem{yfcc100m}
B.~Thomee, D.~A. Shamma, G.~Friedland, B.~Elizalde, K.~Ni, D.~Poland, D.~Borth,
  and L.-J. Li.
\newblock The new data and new challenges in multimedia research.
\newblock {\em arXiv preprint arXiv:1503.01817}, 2015.

\bibitem{attributepops}
N.~Turakhia and D.~Parikh.
\newblock Attribute dominance: What pops out?
\newblock In {\em ICCV}, 2013.

\bibitem{vanDerSluisEtAl06}
I.~van~der Sluis, A.~Gatt, and K.~van Deemter.
\newblock Manual for the tuna corpus: Referring expressions in two domains.
\newblock {\em Technical Report AUCS/TR0705}, 2006.

\bibitem{Viola06multipleinstance}
P.~Viola, J.~C. Platt, and C.~Zhang.
\newblock Multiple instance boosting for object detection.
\newblock In {\em NIPS}, 2006.

\bibitem{WesterbeekEtAl15}
H.~Westerbeek, R.~Koolen, and A.~Maes.
\newblock Stored object knowledge and the production of referring expressions:
  The case of color typicality.
\newblock {\em Frontiers in Psychology}, 6, 2015.

\bibitem{xiao2015learning}
T.~Xiao, T.~Xia, Y.~Yang, C.~Huang, and X.~Wang.
\newblock Learning from massive noisy labeled data for image classification.
\newblock In {\em CVPR}, 2015.

\bibitem{gazedescription}
K.~Yun, Y.~Peng, D.~Samaras, G.~J. Zelinsky, and T.~Berg.
\newblock Studying relationships between human gaze, description, and computer
  vision.
\newblock In {\em CVPR}, 2013.

\bibitem{zhou2004learning}
D.~Zhou, O.~Bousquet, T.~N. Lal, J.~Weston, and B.~Sch{\"o}lkopf.
\newblock Learning with local and global consistency.
\newblock {\em Advances in neural information processing systems},
  16(16):321--328, 2004.

\bibitem{zhu2010semi}
X.~Zhu.
\newblock Semi-supervised learning.
\newblock In {\em Encyclopedia of Machine Learning}. 2010.

\bibitem{zhu2003semi}
X.~Zhu, Z.~Ghahramani, J.~Lafferty, et~al.
\newblock Semi-supervised learning using gaussian fields and harmonic
  functions.
\newblock In {\em ICML}, volume~3, pages 912--919, 2003.

\end{thebibliography}
}

\end{document}